\documentclass[10pt,twocolumn,letterpaper]{article}

\usepackage[pagenumbers]{cvpr} 

\usepackage{graphicx}
\usepackage{amsmath}
\usepackage{amssymb}
\usepackage{booktabs}
\usepackage{multirow}
\usepackage{array}
\usepackage{tabularx}
\usepackage{boldline}

\newcolumntype{Y}{>{\centering\arraybackslash}X}

\usepackage[pagebackref,breaklinks,colorlinks]{hyperref}

\usepackage[capitalize]{cleveref}
\crefname{section}{Sec.}{Secs.}
\Crefname{section}{Section}{Sections}
\Crefname{table}{Table}{Tables}
\crefname{table}{Tab.}{Tabs.}

\def\cvprPaperID{1311} 
\def\confName{CVPR}
\def\confYear{2023}

\begin{document}

\title{Learning Anchor Transformations for 3D Garment Animation}

\author{Fang Zhao$^1$, \ Zekun Li$^1$, \ Shaoli Huang$^{1}$\thanks{Corresponding author.}, \ Junwu Weng$^{1}$, \ Tianfei Zhou$^{2}$, \\ Guo-Sen Xie$^{3}$, \ Jue Wang$^{1}$, \ Ying Shan$^{1}$ \\
	$^1$Tencent AI Lab \ $^2$ETH Zurich \ $^3$Nanjing University of Science and Technology
}
\maketitle

\begin{abstract}
   This paper proposes an \textbf{anchor}-based \textbf{def}ormation model, namely \textbf{AnchorDEF}, to predict 3D garment animation from a body motion sequence. It deforms a garment mesh template by a mixture of rigid transformations with extra nonlinear displacements. A set of anchors around the mesh surface is introduced to guide the learning of rigid transformation matrices. Once the anchor transformations are found, per-vertex nonlinear displacements of the garment template can be regressed in a canonical space, which reduces the complexity of deformation space learning. By explicitly constraining the transformed anchors to satisfy the consistencies of position, normal and direction, the physical meaning of learned anchor transformations in space is guaranteed for better generalization. Furthermore, an adaptive anchor updating is proposed to optimize the anchor position by being aware of local mesh topology for learning representative anchor transformations. Qualitative and quantitative experiments on different types of garments demonstrate that AnchorDEF achieves the state-of-the-art performance on 3D garment deformation prediction in motion, especially for loose-fitting garments.
\end{abstract}

\section{Introduction}
\label{sec:intro}

 \begin{figure}[t]
	\centering
	\includegraphics[width=8cm]{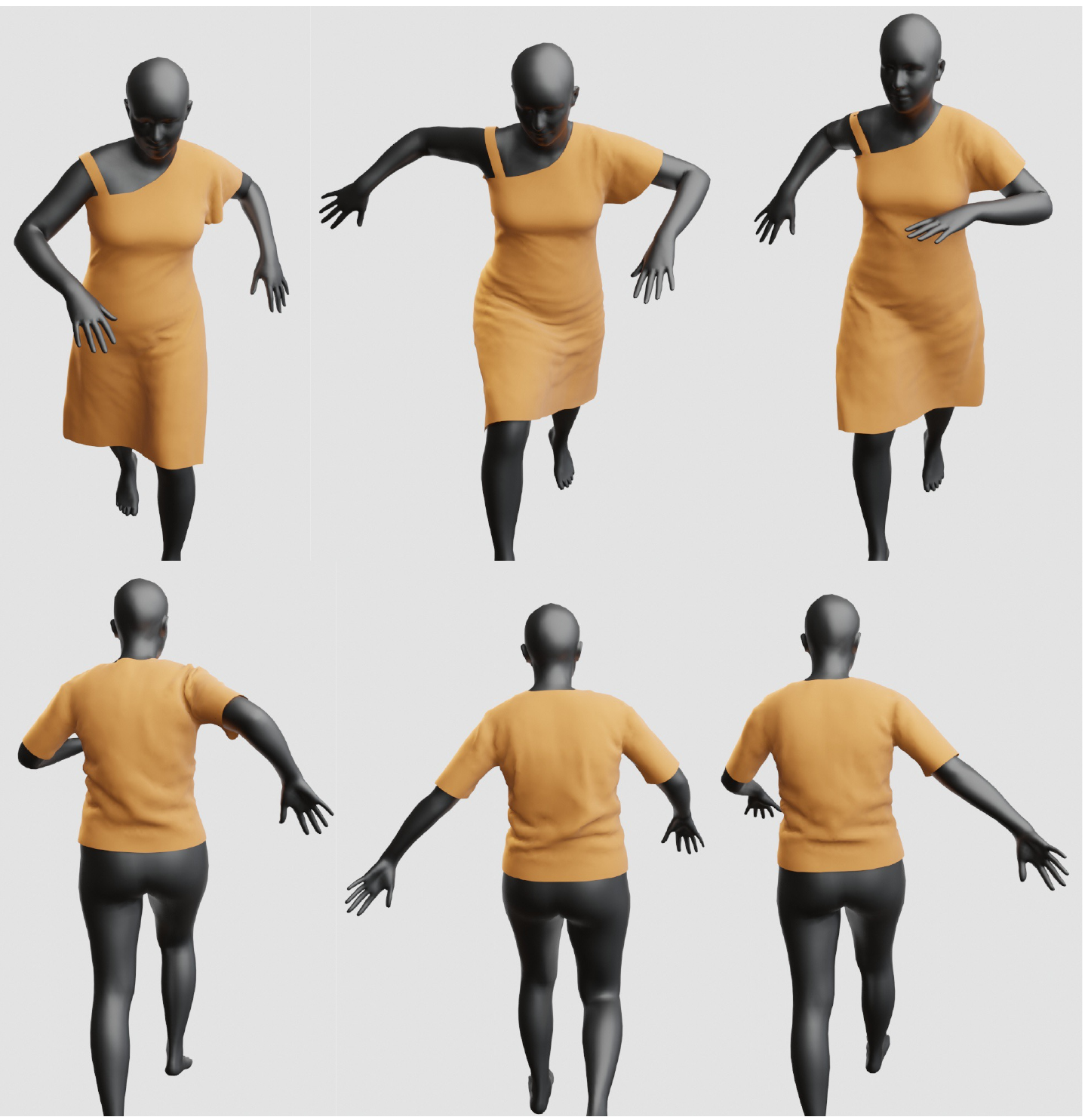}
	\caption{3D garment deformation predicted by the proposed AnchorDEF with body motions. Leveraging the anchor transformations, AnchorDEF is able to realistically deform the garment mesh, especially for loose-fitting garments, \eg, dresses.}\label{fig_intro}
\end{figure}

Animating 3D garments has a wide range of applications in 3D content generation, digital humans, virtual try-on, video games, and so on. Existing pipelines of 3D garment animation usually rely on physics based simulation (PBS), which requires a large amount of computational resources and time costs, particularly for high-quality PBS methods.

Some data-driven or learning-based methods have been proposed to quickly produce 3D garment deformation from static poses or motion sequences with low computational complexity~\cite{feng2010deformation,guan2012drape,pons2017clothcap,lahner2018deepwrinkles,chen2018synthesizing,Gundogdu_2019_ICCV,santesteban2019virtualtryon,patel2020tailornet,zhao2021learning,pfaff2021learning,bertiche2021deepsd,pan2022predicting}. However, many of them attach garment templates to the skeleton of human body for modeling the articulation of garments, which only work with tight garments, \eg, T-shirts and pants, and poorly address loose-fitting ones, \eg, dresses and skirts. In this case, the topology of garments is different from the human body. Therefore, using the skinning blend weights of body yields discontinuities on deformed garment mesh. Several methods~\cite{patel2020tailornet,santesteban2021self} smooth out the blend weights to alleviate the discontinuities but may lose the shape details for large deformations of loose-fitting garments, which do not closely follow body movements.  

To this end, we propose an \textbf{anchor}-based \textbf{def}ormation model, namely \textbf{AnchorDEF}, for predicting 3D garment deformation from a body motion sequence. It leverages a mixture of rigid anchor transformations to represent principle rotations and translations of garments to detach the garment articulation from the body skeleton while preserving the body movement prior, then nonlinear displacements can be regressed relatively easily in a canonical space. As shown in Fig.~\ref{fig_intro}, our method can exploit anchor transformations to realistically deform the garment mesh, especially for loose-fitting garments, \eg, dresses.

Specifically, given a sequence of body motions including poses and translations, we first estimate rigid transformations of a set of anchors around the garment mesh. Using the linear blending skinning (LBS), the garment mesh template is deformed by a weighted combination of the anchor transformations meanwhile per-vertex displacements of the mesh template are regressed to correct artifacts of the blended rigid transformations. To learn physically meaningful anchor transformations for better generalization, we enforce the transformed anchors to maintain consistency with the target's position and normal. In addition, a relative direction constraint is employed to reduce garment-body interpenetrations, which is efficient due to the sparseness of anchors. To make the learned anchor transformations effectively represent the garment deformation, an adaptive anchor updating is further introduced to utilize mesh simplification as supervision to optimize the anchor position. It parameterizes the position with a local attention mask on adjacent mesh vertices and pushes the anchors towards folds and boundaries of garment mesh which usually determine the way of deformation.


The main contributions of our work can be summarized as follows: 1) We propose an anchor-based deformation model which learns a set of anchor transformations and blend weights in a unified framework to represent the deformation of 3D garments, especially for loose-fitting ones. 2) We propose to learn anchor transformations by position and normal consistencies as well as relative direction constraint for better generalization and fewer garment-body interpenetrations. 3) We introduce an adaptive anchor updating with the mesh simplification as supervision to optimize the anchor position for learning representative anchor transformations.

\section{Related Work}
\label{sec:related}

Most garment animation and deformation works can be categorized into physics-based simulations (PBS) and learning-based models.

\noindent\textbf{Physics-based simulations.} PBS methods have been developed for decades and achieved impressive accuracy, leveraging laws of physics and stable accurate solvers~\cite{baraff1998large,harmon2009asynchronous,pcloth20}, collision detection~\cite{muller2015air,provot1997collision,cama16}, and dry frictional contact~\cite{li2022diffcloth,ly2020projective}. But the expensive computational costs, \ie, long simulation time and high-performance devices often are unaffordable in our daily lives.

\noindent\textbf{Learning-based models.} In contrast, learning-based models, which utilize garment animation data attained from PBS or digital scanning to learn the deformation, can often give satisfactory enough animation visual results with less computational costs~\cite{santesteban2019virtualtryon,patel2020tailornet}, regardless of no explicit physical guarantee on accuracy.
Previously, researchers have proposed many techniques depending on the motions of human bodies with linear conditional models~\cite{de2010stable,guan2012drape} or secondary motion graphs~\cite{kim2008drivenshape,kim2013near}, with a constructed database of high-resolution meshes.
Recent works~\cite{patel2020tailornet,santesteban2021self,bertiche2021deepsd} assume that the garment is a submesh of the skinned multi-person linear model (SMPL) and use linear blend skinning (LBS) to transfer body shape to cloth shape.
Beyond learning static results within a single frame, adopting recurrent neural networks to handle the dynamic wrinkles of clothes is widely used~\cite{pan2022predicting,santesteban2019virtualtryon,santesteban2021self}. For example,~\cite{santesteban2019virtualtryon} prepares a database of PBS for dressed characters with some body shapes and motion and uses the database to learn a function of body shape and dynamics for modeling cloth drape and wrinkles. Although learning-based models can be efficient, their performance heavily relies on large amounts of high-quality PBS results or scanning data, which are usually costly~\cite{santesteban2021self,bertiche2020cloth3d,pumarola20193dpeople}.
Thus, some works~\cite{santesteban2022snug,bertiche2020pbns} propose self-supervised training methods to constrain the model with some sort of physical laws to alleviate this problem.
Additionally, to ensure the generalization of simulation, combining several pretrained pivot motion networks can handle different unseen garment styles~\cite{patel2020tailornet} and simulation parameters~\cite{pan2022predicting}. 

But most learning-based methods are limited to tight garments and not suitable for dresses or clothes with more freedom, due to their assumption that garments are complementary to SMPL, \ie, the deformation is closely related to body shape, not the motion. 
To tackle the dynamics, some works propose a neural field to retain high-frequency details~\cite{zhang2021dynamic}, handle challenging areas~\cite{SkiRT3DV2022}, tackle multi-layer garments~\cite{santesteban2021ulnefs}, and warp garments from the canonical pose to the deformation shape~\cite{Chi_2021_ICCV}.
Without utilizing body shape parameters to establish a garment-body aligned model, \cite{wang2019learning} learns an intrinsic cloth shape descriptor embedding space and using different motion parameters to guide garment latent can attain different shape representation. 
To eliminate the alignment between clothes dynamics and body vertices, VirtualBones~\cite{pan2022predicting} extracts a set of rigid bones from the garment animation and uses body motion to predict the bones' transformation to interpolate the garment shape.
Instead of learning based on mesh, \cite{Zakharkin_2021_ICCV} uses a point transformer to predict point clouds of various outfits, for various human poses and shapes, which attains more topological flexibility and the ability to model clothing separately from the body.

\begin{figure*}[t]
	\centering
	\includegraphics[width=17cm]{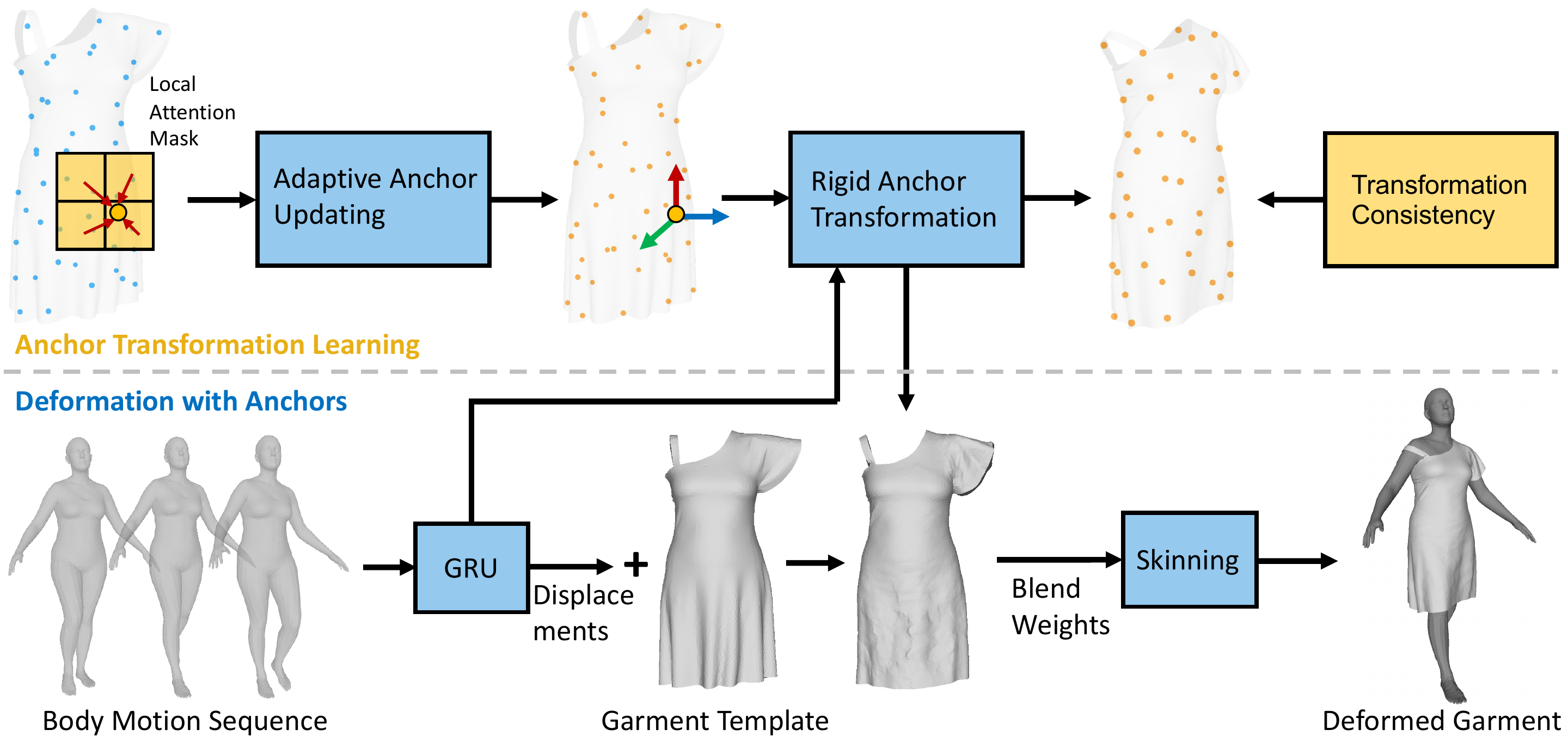}
	\caption{Overview of the proposed AnchorDEF. Given a body motion sequence and a garment mesh template, the garment deformation is predicted by a weighted average of rigid anchor transformations with per-vertex displacements in a canonical space. We explicitly constrain the transformed anchors to satisfy the consistencies of position, normal and direction to guarantee the physical meaning of learned anchor transformations in space for better generalization. Starting from the cluster centers of mesh vertices, the anchor position is updated by a local attention map on adjacent mesh vertices with mesh simplification as supervision. All parameters in our model can be jointly optimized in a unified framework. }\label{fig_overview}
	\vspace{-9pt}
\end{figure*}

VirtualBones~\cite{pan2022predicting} is the most related work to our method, which requires garment skeleton extraction with SSDR~\cite{le2012smooth} as data preprocessing. Different from~\cite{pan2022predicting}, we decompose the garment deformation by a set of adaptive anchors whose rigid transformations and blend weights are learned in a unified framework.
\section{Our Method}
\label{sec:method}

We aim to learn an animated model which is able to predict dynamic 3D garment deformations given a body motion sequence. In this section, we introduce \textbf{AnchorDEF}, an \textbf{anchor}-based \textbf{def}ormation model, which leverages s set of adaptive anchors around mesh surface to deform a garment mesh template by a mixture of rigid anchor transformations with additional per-vertex displacements in a canonical space, where the anchor transformations are learned by the consistencies of position, normal and direction. The overview of AnchorDEF is illustrated in Fig.~\ref{fig_overview}.

\subsection{Deformation with Anchors}

3D animated models of human body, \eg, SMPL~\cite{loper2015smpl}, usually deform a body mesh template $\mathbf{T}$ with blendshapes and skinning, which can be defined as
\begin{equation}\label{eq_anim}
	V=W(\mathbf{T}+B( \theta), \mathbf{J}, \theta, \mathbf{w}),
\end{equation}
where $W$ is a skinning function, \eg, linear blend skinning, with a skeleton $\mathbf{J}$, pose parameters $\theta$ and blend weights $\mathbf{w}$, $V$ is the posed vertices and $B$ is the blendshape corresponding to the body pose $\theta$. Since the garment is a kind of soft body and it is hard to directly predict the coordinates of each vertex from a body motion sequence, here we also consider to decompose the garment deformation into two parts including a weighed average of rigid transformations and nonlinear displacements. 

Most data-driven garment models directly associate garments with the skeleton of human body~\cite{santesteban2019virtualtryon,patel2020tailornet,bertiche2021deepsd}, \ie, they apply the skinning function to animate the garment by directly using the blend weights of body mesh vertices in the neighborhood, where it is assumed that garments closely follow the body movement. However, for loose-fitting garments, the clothing deformation usually contains more complex nonrigid effects compared to human body which can be seen as a rigid body. In this case, garments behave significantly differently from the body. Thus, it is unreasonable to attach the garment to the body skeleton. 

To address this issue, we adopt a set of anchors $\{p_n\}$ around the mesh surface to guide the garment deformation, which detaches the garment from the body skeleton while preserving the body movement prior. Given a body motion sequence consisting of poses and translations, rigid anchor transformations including rotations $\{R_{n}\}$ and translations $\{T_{n}\}$ are estimated firstly by using Gated Recurrent Units (GRU)~\cite{cho2014properties} (see the supplementary material for the network structure). Then, the $m$-th deformed garment vertex $v_m$ can be represented by a weighted average of anchor transformations:
\begin{equation}\label{eq_wrt}
	v_{m}=\sum_{n=1}^{N} w_{mn}({R_n}v^c_m+T_n),  w_{mn}\ge 0 \text{ and } \sum_{n=1}^{N} w_{mn} =1,
\end{equation}
where $v^c_m$ is the $m$-th vertex of garment mesh template, $w_{mn}$ is the blend weight of $v_m$ regarding to the $n$-th anchor and $N$ is the number of anchors. To determine the anchor-vertex relationship and maintain the sparsity of $\{w_n\}$, we initialize the anchors as the clustering centers of vertices of the mesh template. Then, each vertex is only influenced by the blend weights of its $N'$ ($N'\ll N$) nearest anchors and the others remain zeros during training. Compared to the direct prediction of deformed mesh vertex coordinates, the rigid anchor transformations provide more generalized representations of complex deformations due to fewer parameters required to be learned for the prediction.

Two strategies of anchor transformation configuration are designed for loose and tight types of garments. For the loose type, to utilize the prior of the global human body orientation $R_{global}$, we predict transformations relative to the root joint $J_{root}$ of the body and the anchor transformation can be defined as:
\begin{align}\label{eq_loose}
	R^{loose}_n &= R_{global}R_n([\theta,t]), \nonumber \\
	T^{loose}_n &= R_{global}(T_n([\theta,t]) - R_n([\theta,t])J_{root}) + J_{root},
\end{align}
where $\theta$ and $t$ are the body pose and translation, respectively. For the tight type, the garment mostly follows the body movement. Thus, we predict an incremental change on the basis of body vertex rotation $R_{body}$ and offset $T_{body}$:
\begin{align}\label{eq_tight}
	R^{tight}_n &= R_{body}R_n([\theta,t]), \nonumber \\
	T^{tight}_n &= R_{body}T_n([\theta,t]) + T_{body},
\end{align}
where $R_{body}$ and $T_{body}$ is the transformation of the body mesh vertex closest to the anchor $p_n$. Furthermore, we can combine these two strategies for garments with both tight and loose parts, \eg, dresses, by classifying which part the anchor belongs to according to its distance to body joints.

Once the primary rotations and translations in the garment deformation caused by the articulation are expressed in Eq.~\eqref{eq_wrt}, the nonlinear residual deformation can be disentangled with the garment articulation and defined as per-vertex displacements $D([\theta,t])$ in an unposed canonical space. Utilizing the skinning function in Eq.~\eqref{eq_anim}, the overall garment deformation is given by
\begin{equation}\label{eq_defom}
	V_{pred}=W(\mathbf{T}_g+D([\theta,t]),G([\theta,t]),\mathbf{w}_p),
\end{equation}
where $\mathbf{T}_g$ is the garment mesh template, $\mathbf{w}_p$ is the anchor-vertex blend weights and $G([\theta,t])$ is the anchor transformations predicted with the body motion. We argue that such decomposition by anchors extremely reduces the complexity of pose-dependent garment deformation prediction because large deformations have been handled by transformations of sparse anchors and only small displacements need to be regressed in the canonical space.


\subsection{Learning Anchor Transformations}

Different from~\cite{pan2022predicting} which resorts to the external skinning decomposition algorithm, \ie, SSDR~\cite{le2012smooth}, to extract the garment skeleton from smoothed mesh sequences, here we jointly learn rigid transformations, blend weights and positions of anchors in a unified framework. 

We first apply an L2 loss term on the predicted mesh vertices to recover the shape of deformed mesh:
\begin{equation}\label{eq_l2}
	\mathcal{L}_{rec} = ||V_{pred}-V_{gt}||^2_2,
\end{equation}
which minimizes the Euclidean distance between the predicted vertices and the ground-truth deformed vertices. Then a loss term is used to keep the similarity of the graph Laplacians between two meshes:
\begin{equation}\label{eq_lap}
	\mathcal{L}_{lap} = ||\bigtriangleup V_{pred}-\bigtriangleup V_{gt}||^2_2.
\end{equation}
A collision term~\cite{pan2022predicting} is also added to avoid garment-body collisions. Thus, the loss on the vertex can be given by
\begin{equation}\label{eq_loss_vert}
	\mathcal{L}_{vert} = \mathcal{L}_{rec} + \beta_{1}\mathcal{L} _{lap} + \beta_{2}\mathcal{L}_{collision},
\end{equation}
where $ \beta_{1}$ and $\beta_{2}$ are trade-off coefficients for controlling the degree of smoothing and collision, respectively.

 \begin{figure}[t]
	\centering
	\includegraphics[width=8.0cm]{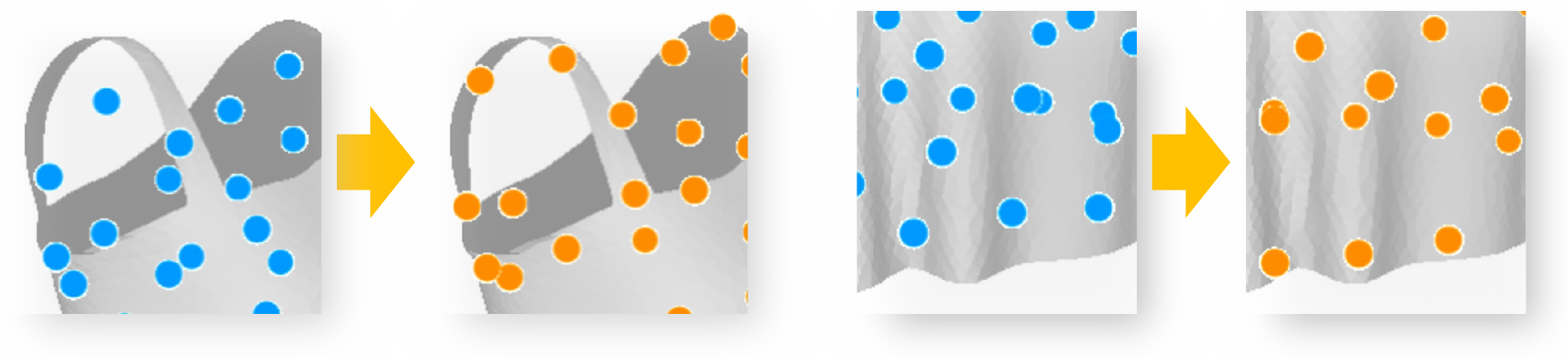}
	\caption{Examples of adaptive anchor updating by using the local attention mask with the mesh simplification as supervision. The blue points are anchors initialized by \textit{k}-means and the orange ones are the updated anchors.
	}\label{fig_update}
\end{figure}

\begin{figure*}[t]
	\centering
	\includegraphics[width=17.0cm]{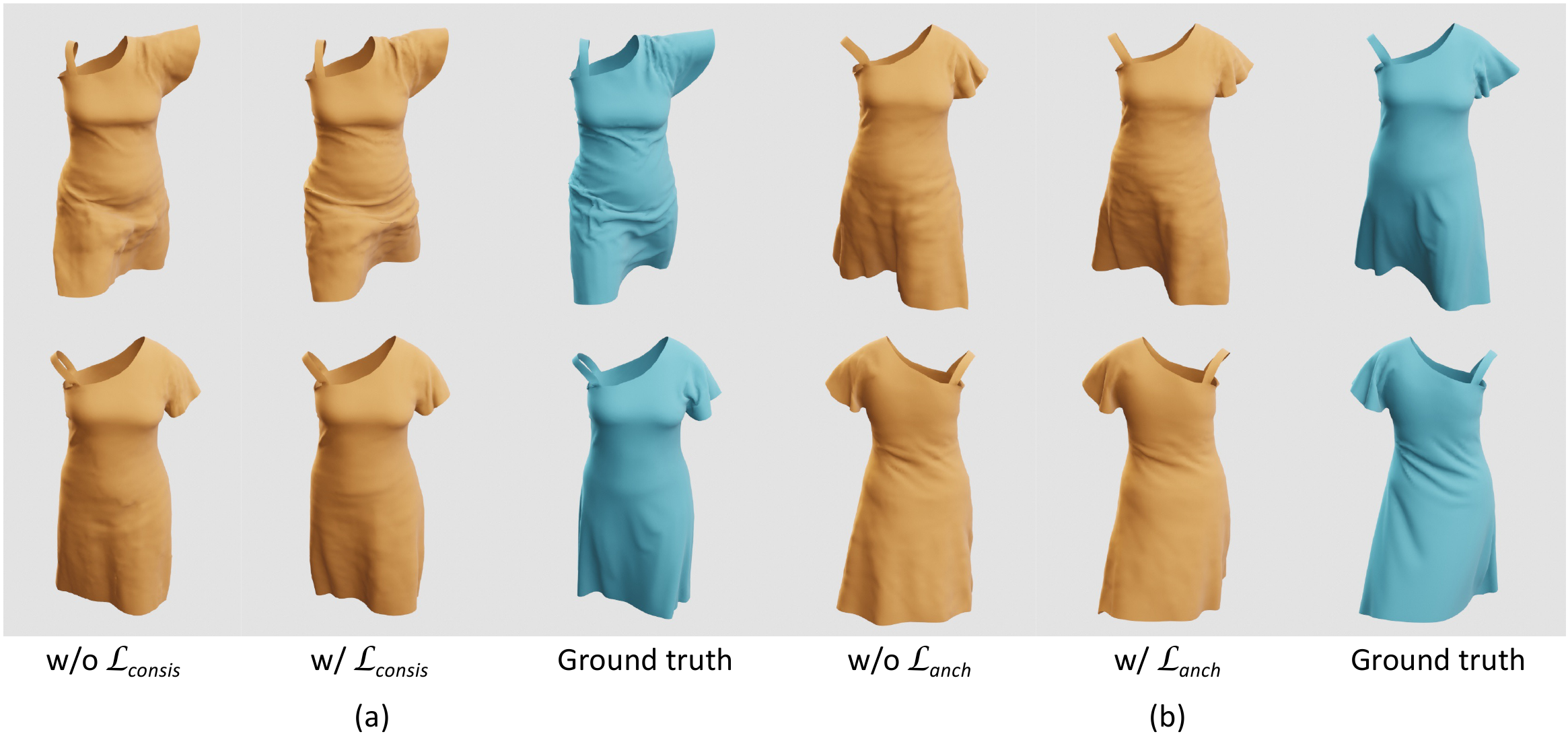}
	\vspace{-8pt}
	\caption{Qualitative comparison with and without different components of AnchorDEF including (a) anchor transformation consistency $\mathcal{L}_{consis}$ and (b) adaptive anchor updating $\mathcal{L}_{anch}$. Without $\mathcal{L}_{consis}$, the model cannot accurately predict the shape of deformed mesh and loses some surface geometry details. Without $\mathcal{L}_{anch}$, the model is unable to produce realistic garment deformations due to the under-expressed anchor transformations.}\label{fig_consis}
\end{figure*}

However, only optimizing the vertex position cannot guarantee the physical meaning of learned anchor transformations in space, which leads to poor generalization. Thus, we explicitly constrain the transformed anchors to be located around the deformed mesh surface. To this end, the position of anchor $p$ is recalculated by its adjacent mesh vertices:
\begin{equation}\label{eq_anchor_pos}
	p = \frac{1}{K} \sum_{v_k^c\in K(p, \mathbf{T}_g)} v_k^c,
\end{equation}
where $K(p, \mathbf{T}_g)$ is the $K$ nearest vertices of $p$ on the mesh template. Then, the target point $p_{tgt}$ of transformed anchor $p_G$ can be defined with the same set of vertices $K(p, \mathbf{T}_g)$ on the deformed mesh and a transformation consistency loss is defined as
\begin{equation}\label{eq_loss_consis}
	\mathcal{L}_{consis} = ||p_G-p_{tgt}||^2_2 + \gamma||\mathbf{n}(p_G)-\mathbf{n}(p_{tgt})||^2_2,
\end{equation}
where $\mathbf{n}(.)$ is the unit normal of point and $\gamma$ is a weight factor. By encouraging the transformed anchor $p_G$ to be similar to $p_{tgt}$ in position and normal, the learned anchor transformation indicates the mean transformation of $K$ nearest vertices of $p$. It can be viewed as a regularization for the transformation magnitude and direction to discover essential transformations in deformation.


We further enforce the direction from $p_{tgt}$ to $p_G$ to be consistent with the direction of $\mathbf{n}(p_{tgt})$ by minimizing the following penalty term:
\begin{equation}\label{eq_pc}
	\mathcal{L}_{dir}=1 - \cos(p_G-p_{tgt}, \mathbf{n}(p_{tgt})),
\end{equation}
where $\cos(\cdot, \cdot)$ is the cosine similarity. Intuitively, this penalty term makes $p_G$ tend to be outside of the plane formed by $p_{tgt}$ to reduce the garment-body interpenetration.


\noindent\textbf{Adaptive Anchor Updating.} It is still not enough to have a physically meaningful anchor transformation, we also need an expressive transformation in geometry to better represent the garment deformation. We observe that mesh regions with folds and boundaries usually play an important role on showing garment shape characteristics and contain more deformation information compared with smooth regions. 

To make the anchors reflect such intrinsic structure of garment shape, we propose to employ mesh simplification to provide a signal of supervision for finding such representative anchors. Here Quadric Error Metric (QEM)~\cite{garland1997surface} is used to obtain simplified mesh vertices by collapsing edges while preserving the garment topology. Because there is no one-to-one correspondence between the anchors $\{p_n\}$ and the simplified vertices $\{{\tilde v}_m\}$, the Chamfer distance is adopted as the optimization objective:
\begin{equation}\label{eq_loss_anch}
	{\mathcal{L}_{anch}} = \sum\limits_{p} {\mathop {\min }\limits_{{\tilde v}} ||{p} - {{\tilde v}}||_2^2}  + \sum\limits_{{\tilde v}} {\mathop {\min }\limits_{p} ||{p} - {{\tilde v}}||_2^2}.
\end{equation}
To be able to update the anchor position, we rewrite Eq.~\eqref{eq_anchor_pos} using learnable weights $\alpha$ as
\begin{equation}\label{eq_anchor_w}
	p = \frac{1}{\sum_{j=1}^{K}e^{\alpha_j}} \sum_{k=1}^{K} e^{\alpha_k} v_k^c.
\end{equation}
Actually, this equals to learning a local attention mask for the anchor to determine how the adjacent vertices contribute to its position. Fig.~\ref{fig_update} shows that by optimizing the weights $\alpha$ with supervision of the simplified vertices, the anchors are pushed to inflection points of folds and boundaries of the mesh and thus able to better represent the deformed mesh vertices by interpolating their transformation matrices. Note that we optimize the anchor position on the entire dataset not only on the mesh template because some smooth regions of the template may create new folds after posing. 


On the other hand, if we reformulate the transformed anchor $p_G$ with Eq.~\eqref{eq_anchor_w} as:
\begin{equation}\label{eq_anchor_rew}
	p_G = Rp+T = \frac{1}{\sum_{j=1}^{K}e^{\alpha_j}} \sum_{k=1}^{K} e^{\alpha_k} (Rv_k^c + T),
\end{equation}
the process of transformation learning for the anchor, \ie., Eq.~\eqref{eq_loss_consis}, can be interpreted as a way to find an appropriate transformation for its adjacent mesh vertices by the importance which is reflected in the weights learned from the mesh simplification. The larger the weight, the greater the impact of the vertex transformation on the anchor transformation learning.

Lastly, combining Eq.~\eqref{eq_loss_vert}, \cref{eq_loss_consis,eq_pc,eq_loss_anch}, the overall objective function for our model is given by:
\begin{equation}\label{loss_final}
	\mathcal{L} = \mathcal{L} _{vert} + \lambda_{1}\mathcal{L} _{consis} + \lambda_{2}\mathcal{L}_{dir} + \lambda_{3}\mathcal{L} _{anch},
\end{equation}
where $\lambda$s are loss weights.

\section{Experiments}
We evaluate the proposed AnchorDEF on the VTO~\cite{santesteban2019virtualtryon} dataset containing two types of garments, \ie, dress and T-shirt. Following~\cite{pan2022predicting},  RMSE (Root Mean Squared Error), Hausdorff distance and STED (Spatio-Temporal Edge Difference)~\cite{vasa2010perception} are adopted to assess the quality of deformed meshes. The first two measure the distance between predicted and ground truth meshes and STED reflects the difference between their dynamics. We also compare AnchorDEF against other 3D garment deformation methods~\cite{pan2022predicting,patel2020tailornet}. 

\noindent\textbf{Implementation Details.} To capture the motion dynamics, a network with 2 GRU layers of size 256 is used as the body motion encoder. Two 2-layered MLPs of size 256 are used to regress the anchor transformations and per-vertex displacements, respectively. The number of anchors associated with each mesh vertex is set to 8, which can obtain the desired smooth surface, and the number of neighboring vertices for computing the anchor position is set to 128. For the garment sequences of VTO, 4 clips are randomly selected for testing and the rest is used as the training set, and the length of training sequence is 30. Please refer to the supplemental materials for more details on training and inference.
\begin{figure}[t]
	\centering
	\includegraphics[width=8.0cm]{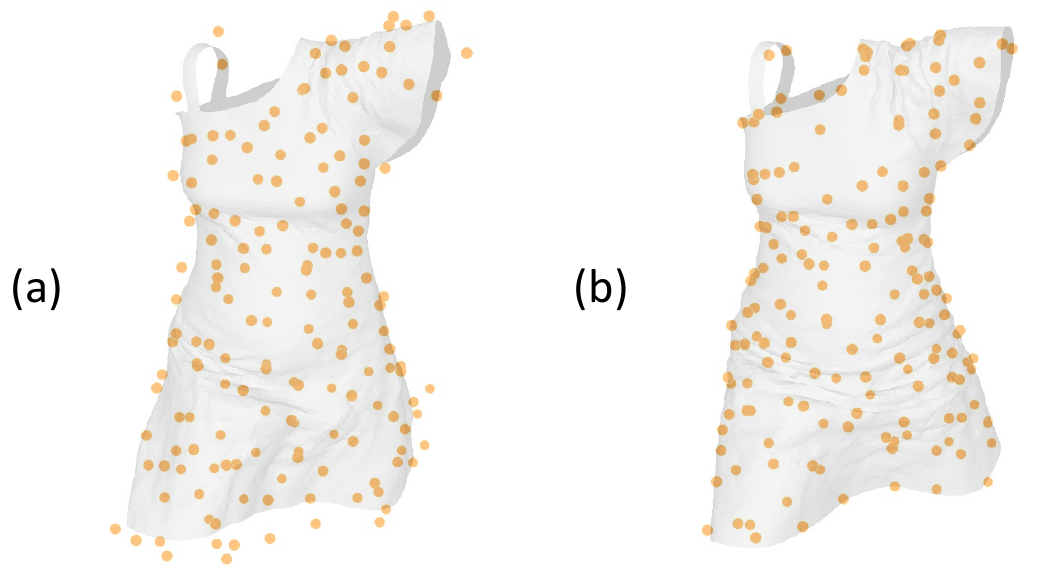}
	\caption{Visualization of anchors transformed by (a) only optimizing the mesh vertices during training and (b) AnchorDEF.}\label{fig_anch_vis}
\end{figure}

\begin{table}[t]
	\centering
	\begin{tabularx}{0.47\textwidth}{lYYY@{}}
		\hlineB{2.5}
		Methods &  w/o $L_{consis}$ & w/o $L_{anch}$  & AnchorDEF \\
		\hline
		RMSE $\downarrow$ & 17.21 & 16.49  & \textbf{16.05} \\
		Hausdorff $\downarrow$ & 75.01 & 74.53  & \textbf{74.20} \\
		STED $\downarrow$ & 0.0595 & 0.0526  & \textbf{0.0493} \\
		\hlineB{2.5}
	\end{tabularx}
	\caption{Quantitative results obtained by different components of AnchorDEF, \ie, without anchor transformation consistency (w/o $\mathcal{L}_{cosis}$ ), without adaptive anchor updating (w/o $\mathcal{L}_{anch}$) and the full model (AnchorDEF).}
	\label{tab_ablation}
\end{table}

\begin{figure}[t]
	\centering
	\includegraphics[width=8.3cm]{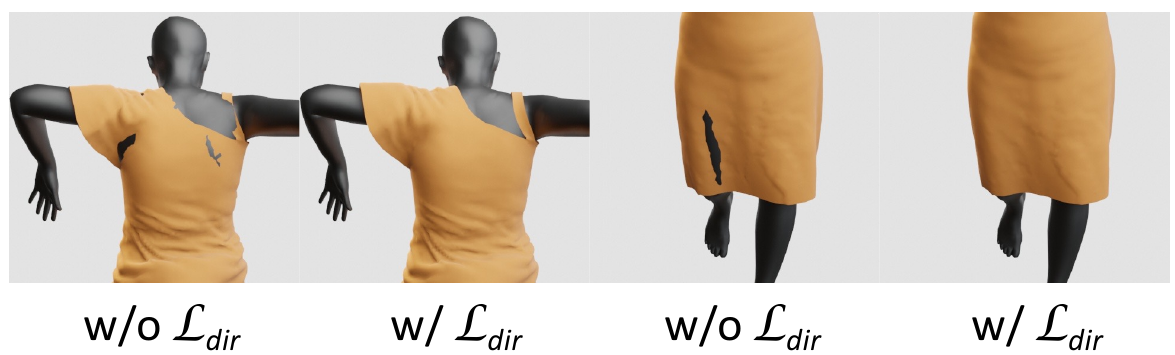}
	\caption{Qualitative comparison with and without the penalty $\mathcal{L}_{dir}$  on the direction from the target anchor to the transformed one.}\label{fig_pc}
\end{figure}

\begin{figure}[t]
	\centering
	\includegraphics[width=8.3cm]{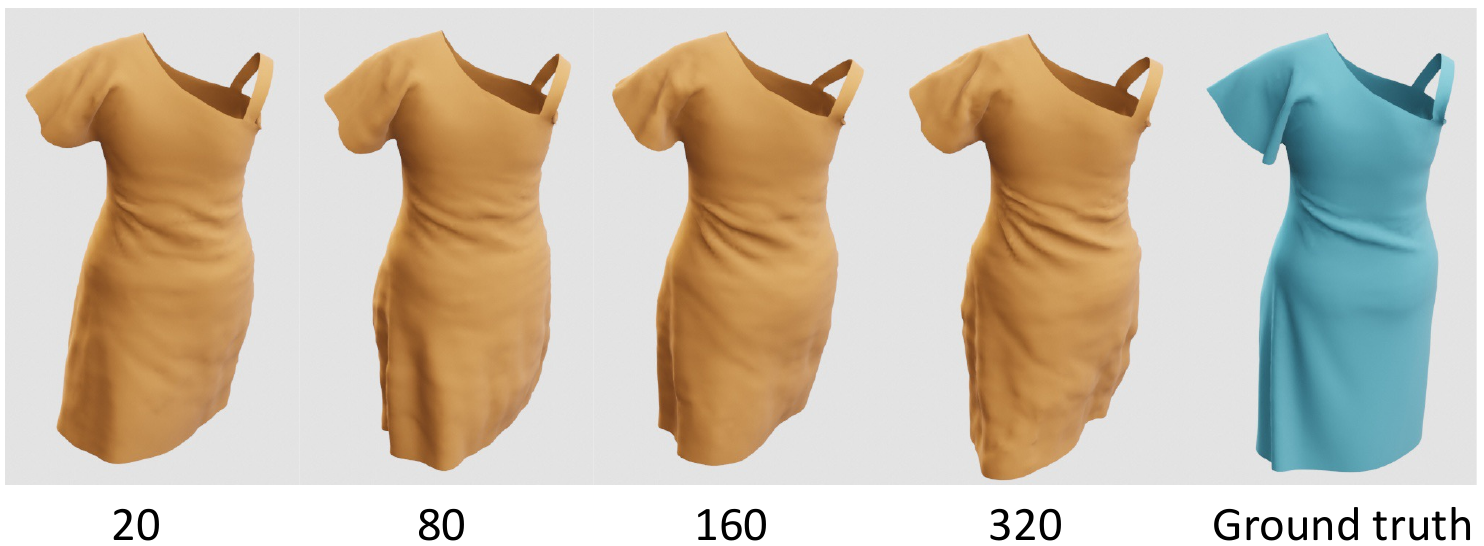}
	\caption{Qualitative comparison of using different numbers of anchors. Using 160 anchors produces finer and more accurate deformation compared with other anchor numbers. }\label{fig_anch_num}
\end{figure}

\begin{table}[t]
	\centering
	\begin{tabularx}{0.47\textwidth}{lYYYY@{}}
		\hlineB{2.5}
		Num. Anchor & 20  & 80 & 160 & 320 \\
		\hline
		RMSE $\downarrow$ &  17.78 & 16.82  & \textbf{16.05} & 17.52  \\
		Hausdorff $\downarrow$ &  75.30 &  74.71  & \textbf{74.20}  & 75.13  \\
		STED $\downarrow$ &  0.0620  &  0.0539 & \textbf{0.0493}  & 0.0625  \\
		\hlineB{2.5}
	\end{tabularx}
	\caption{Quantitative results of using different numbers of anchors.}
	\label{tab_num_anch}
	\vspace{-8pt}
\end{table}

\subsection{Ablation Study}

We validate the effectiveness of main components in our proposed AnchorDEF. We firstly assess the importance of anchor transformation consistency, \ie, $\mathcal{L}_{consis}$ in Eq.~\eqref{eq_loss_consis}. As reported in Table~\ref{tab_ablation}, keeping the consistency of learned anchor transformations is able to improve all metrics due to better generalization. Fig.~\ref{fig_consis} (a) illustrates the qualitative comparison. One can observe that without $\mathcal{L}_{consis}$ the model cannot accurately predict the shape of deformed mesh and loses some surface geometry details. 

 \begin{figure*}[t]
	\centering
	\includegraphics[width=17.0cm]{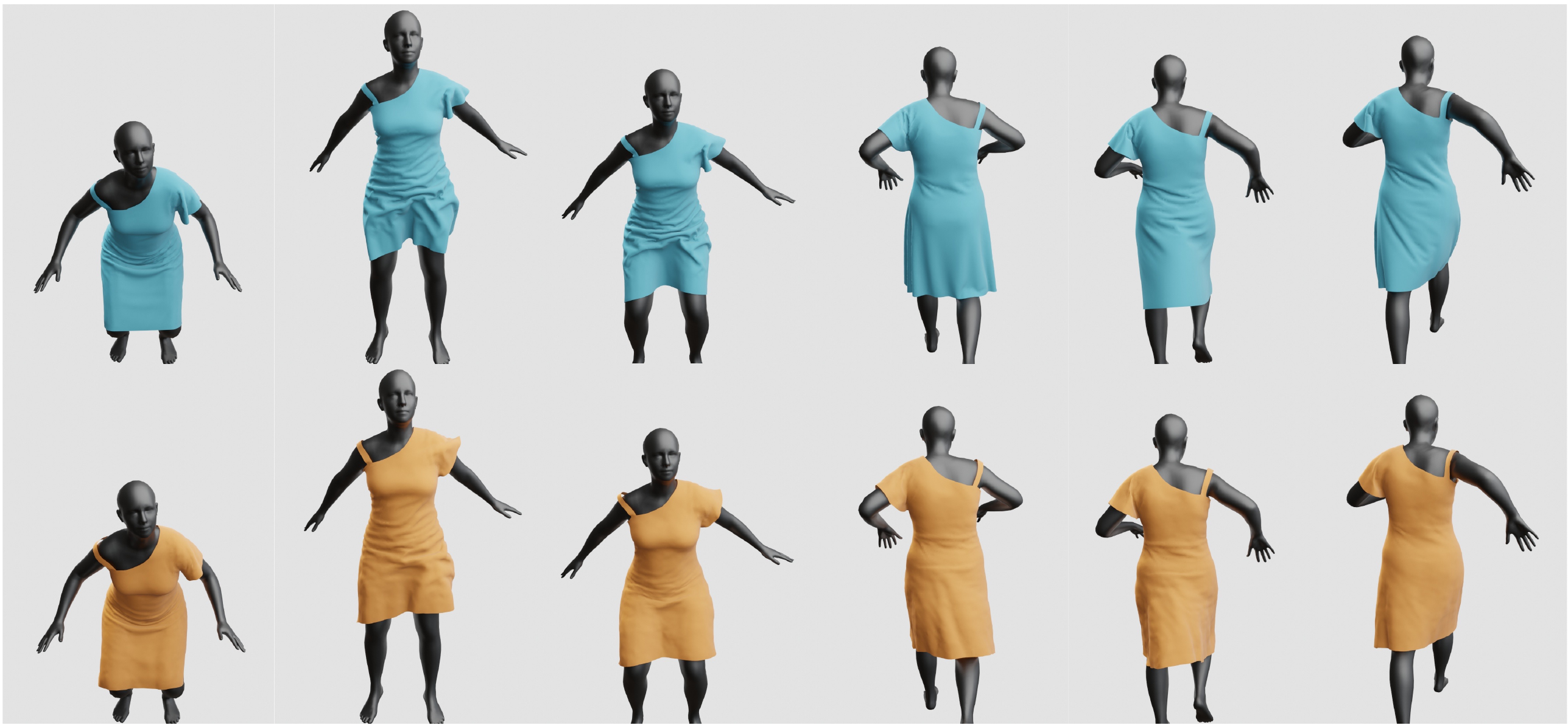}
	\caption{Examples of our AnchorDEF for dynamic garment deformation in motion. The first row is the ground truth generated by physics-based simulation~\cite{narain2012adaptive}. The second row is the results obtained by AnchorDEF. Our method can produce natural and realistic clothing dynamics, \eg, wrinkles from jumping and the swing of the dress caused by running.}\label{fig_unseen}
	\vspace{-5pt}
\end{figure*}

To evaluate the influence of adaptive anchor updating, Table~\ref{tab_ablation} also reports the results without using $\mathcal{L}_{anch}$ in Eq.~\eqref{eq_loss_anch}. The performance drops only using the clustering centers of \textit{k}-means as the anchors. The reason is that compared with the clustering, the supervision with mesh simplification can provide key topology information, \eg, folds and boundaries, which is useful for learning representative anchor transformations. The qualitative results are shown in Fig.~\ref{fig_consis} (b). It can be observed that without $\mathcal{L}_{anch}$ the model is unable to produce realistic garment deformations due to the under-expressed anchor transformations.

Fig.~\ref{fig_anch_vis} (a) and (b) visualizes the anchors transformed by only optimizing the mesh vertices during training and AnchorDEF, respectively. For Fig.~\ref{fig_anch_vis} (a), without the consistency constraint for anchor transformation, many of the transformed anchors are not on the deformed mesh surface, which is lack of clear physical meaning in space and easy to overfit. In contrast, AnchorDEF enforces the anchors to be transformed onto the deformed mesh surface, which prevents overfitting to a certain extent, thus has better generalization when testing. Furthermore, as shown in Fig.~\ref{fig_anch_vis} (b), the transformed anchors are mainly distributed at the folds and boundaries of the mesh because of our adaptive anchor updating according to simplified mesh vertices, and thus capable of better expressing the deformation of mesh vertices, in particular those vertices far from them. In this way, the complexity of garment deformation space is reduced more effectively and more mesh shape details can be preserved.

Then we investigate the impact of the penalty $\mathcal{L}_{dir}$ (Eq.~\eqref{eq_pc}) on the direction from the target anchor to the transformed one. As illustrated in Fig.~\ref{fig_pc}, with $\mathcal{L}_{dir}$ the model can reduce the garment-body interpenetration. By enforcing the transformed anchors to be on the positive direction of target normals, the garment mesh vertices influenced by these anchors can be effectively pushed to the outside of the body surface while keeping the deformation accuracy due to moving along the direction of target normals.

We further assess the influence of the anchor number. Table~\ref{tab_num_anch} reports the results obtained by using 20, 80, 160 and 320 anchors, respectively. One can see that using more anchors to guide the mesh deformation obtains better results on all three metrics because the decomposition for the mesh deformation by using the rigid anchor transformations makes the nonlinear deformation can be more easily predicted as the small displacements in the canonical space without considering large rotations and translations caused by the body pose which has been handled by the anchor transformations. However, the performance drops as the number of anchors continues to increases because too many transformation parameters lead to unstable forecasts during testing. Fig.~\ref{fig_anch_num} shows the qualitative comparison. Using 160 anchors produces finer and more accurate deformation compared with other anchor numbers. Using 20 anchors only recover the contour of the deformed mesh, and using 80 and 320 anchors fail to obtain realistic deformation.

Fig.~\ref{fig_unseen} shows some examples of our AnchorDEF for dynamic garment deformation in motion. Given a body motion sequence, our method can produce natural and realistic clothing dynamics, \eg, wrinkles from jumping and the swing of the dress caused by running.

\begin{table*}[t]
	\centering
	\begin{tabularx}{0.9\textwidth}{lYYYYYY@{}}
		\hlineB{2.5}
		\multirow{2}{*}{Methods} & \multicolumn{3}{c}{Dress} & \multicolumn{3}{c}{T-shirt}  \\
		\cline{2-4} \cline{5-7}
		& RMSE $\downarrow$ & Hausdorff $\downarrow$ & STED $\downarrow$ & RMSE $\downarrow$ & Hausdorff $\downarrow$ & STED $\downarrow$ \\
		\hline
		TailorNet~\cite{patel2020tailornet} & 22.95  & 76.80 & 0.0757 & 9.90 & 27.02 & 0.0418 \\
		VirtualBones~\cite{pan2022predicting} & 19.91 & 83.39 & 0.0722 & 10.52 & 31.51 & 0.0452 \\
		AnchorDEF & \textbf{16.05} & \textbf{74.20} & \textbf{0.0493} & \textbf{6.25} & \textbf{26.31} & \textbf{0.0262} \\
		\hlineB{2.5}
	\end{tabularx}
	\vspace{-3pt}
	\caption{Quantitative results of 3D garment deformation methods for different types of garments.}
	\label{tab_sota}
\end{table*}

 \begin{figure*}[t]
	\centering
	\includegraphics[width=17.0cm]{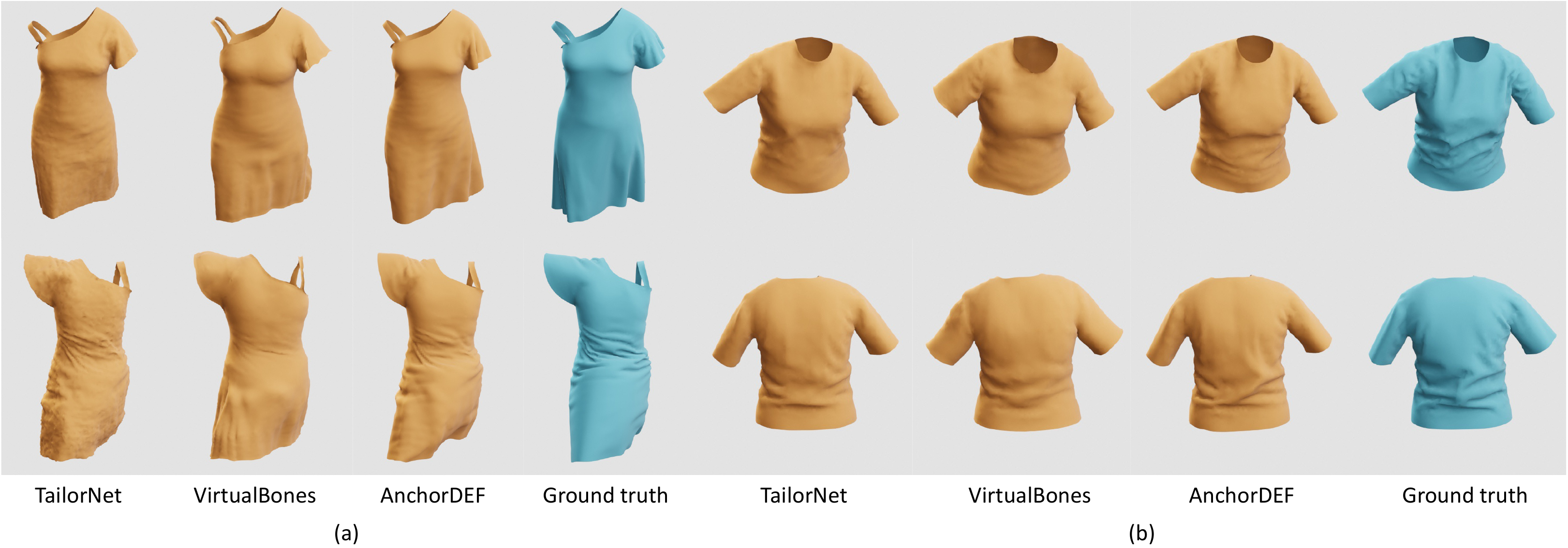}
	\vspace{-6pt}
	\caption{Qualitative comparison of 3D garment deformation methods for different types of garments, including TailorNet~\cite{patel2020tailornet}, VirtualBones~\cite{pan2022predicting} and Our AnchorDEF. With the learned anchor transformations, our method not only recovers large deformations of the dress but also produces fine surface details on the T-shirt.}\label{fig_sota}
\end{figure*}

\subsection{Comparison with Related Methods}

We compare our method against related 3D garment deformation methods, including TailorNet~\cite{patel2020tailornet} and VirtualBones~\cite{pan2022predicting} on different types of garments. As reported in Table~\ref{tab_sota}, our AnchorDEF achieves the best results in all cases. Fig.~\ref{fig_sota} presents the qualitative comparison. TailorNet~\cite{patel2020tailornet} recovers folds on the T-shirt, but it fails to address the deformation of the dress which has different topology from human body. By interpolating the skinning blend weights, it alleviates the discontinuity of blend weights borrowed from the SMPL~\cite{loper2015smpl} body. However, this makes TailorNet only produce the outline of shapes with some folds missing and artifacts for large deformations of loose-fitting garments. VirtualBones~\cite{pan2022predicting} can estimate large deformations of the dress by using SSDR~\cite{le2012smooth} to extract virtual bones of garments. However, it tends to keep the original folds of the dress template on the deformed mesh and cannot estimate well new folds generated by body motions on the T-shirt. VirtualBones needs SSDR to extract the garment skeleton and gets blend weights of bones in advance and fixes the bone position and blend weights during network training. Moreover, SSDR is performed on the low-frequency mesh sequences processed by Laplacian smoothing because it fails to address high-frequency garment deformations. Thus, the learned virtual bones are not necessarily optimal. In contrast, by jointly learning the rigid transformations, blend weights and positions of anchors, our AnchorDEF not only recovers large deformations of the dress but also produces fine surface details on the T-shirt. More qualitative results can be found in the supplemental materials.

\section{Conclusions}

We present AnchorDEF, an anchor based deformation model, to predict 3D garment animation given a body motion sequence. With a set of anchors around the mesh surface, the garment deformation can be decomposed into a mixture of rigid anchor transformations and extra per-vertex displacements in a canonical space. The transformation consistency learning is introduced to guarantee the physical meaning of learned anchor transformations in space. Taking mesh simplification as supervision, we further optimize the anchor position for learning representative anchor transformations by making the anchor aware of local mesh topology with an attention mask. Extensive experiments on different types of garments demonstrate that AnchorDEF can effectively predict 3D garment deformation in motion for both tight and loose-fitting garments.

\noindent\textbf{Limitations.} Compared with the PBS methods, our method still lacks fine-level dynamics. Better exploiting the temporal information for the network architecture and supervision signals is a good direction for future work.

\noindent\textbf{Acknowledgments.} This work was partly supported by the National Natural Science Foundation of China (Nos. 62276134).

{\small
\bibliographystyle{ieee_fullname}
\bibliography{zfbib_3d_cvpr2023}
}

\clearpage

\section*{Supplemental Materials}

In the supplemental materials, we present additional details on training and inference procedures and more qualitative results and comparisons.

\section*{A. Training and Inference}
\label{model}

For training, we adopt the Adam optimizer~\cite{kingma2014adam} with an initial learning rate of 1e-3. The batch size is 8 and the number of epochs is 50. The learning rate is lowered to 1e-4 after 30 epochs. Empirically the weights ${\lambda}_1$, ${\lambda}_2$ and ${\lambda}_3$ of the overall objective function are set to 1, 0.01 and 100, respectively. The weight factor $\gamma$ in the transformation consistency loss is set to 0.1. The Laplacian and collision coefficients in the vertex loss are set to 0.2 and 1, respectively. At the beginning of training, we optimize the objective function without the collision term and the direction penalty term and add these two terms  in the last 10 epochs while recalculating the anchor-vertex relationships according to the new anchor positions.

\begin{figure}[t]
	\centering
	\includegraphics[width=8.0cm]{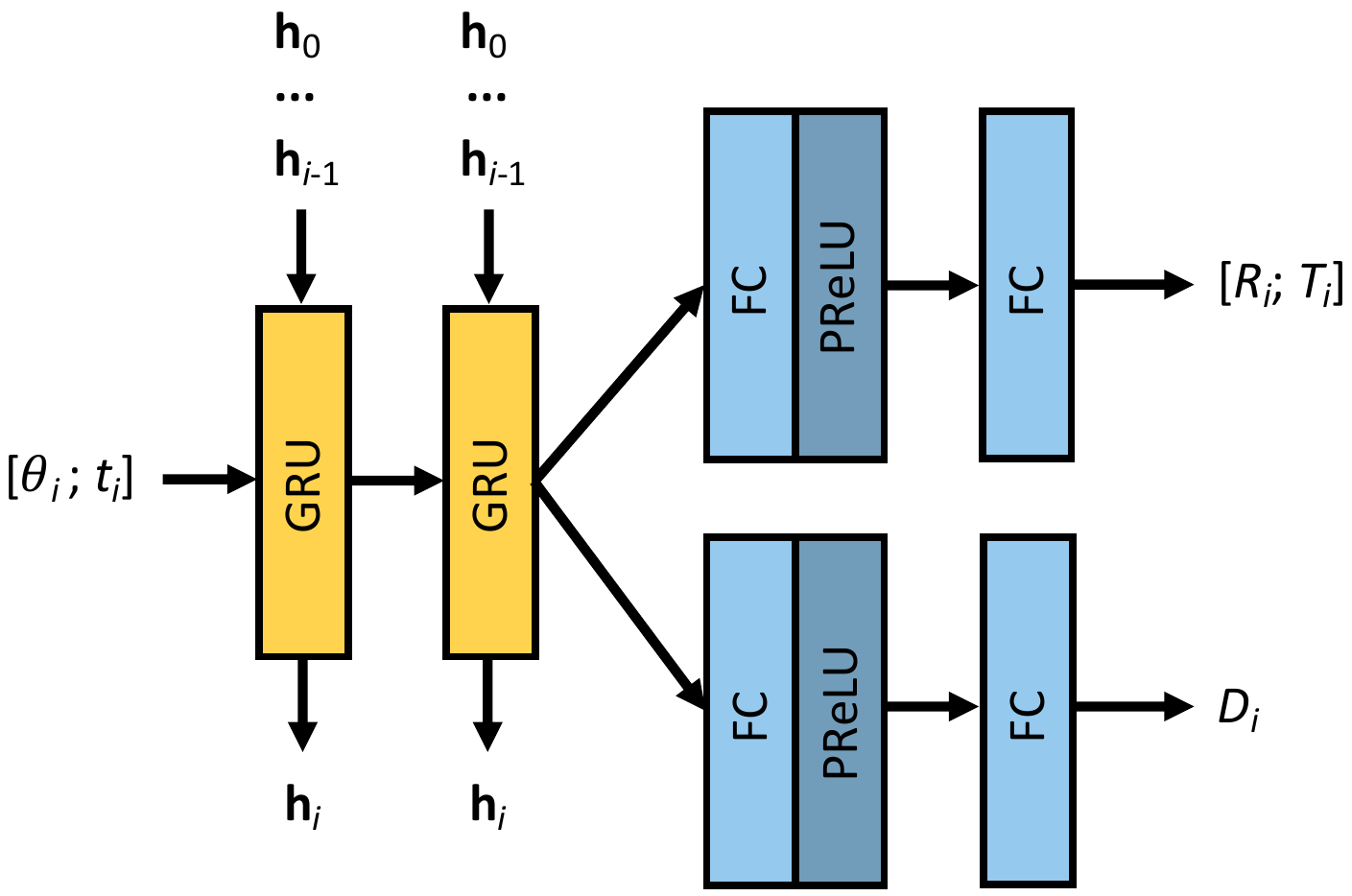}
	\vspace{-3pt}
	\caption{Structure of the network for estimating anchor rotations and translations $[R_i; T_i]$ and per-vertex displacements $D_i$ in the canonical space.}\label{fig_gru}
	\vspace{-3pt}
\end{figure}

Fig.~\ref{fig_gru} shows the structure of the network for estimating anchor transformations and per-vertex displacements in the canonical space. For the \textit{i}-th frame, the body pose ${\theta}_i$ and translation $t_i$ are concatenated and fed into the GRU layers. Anchor rotations and translations $[R_i; T_i]$ and per-vertex displacements $D_i$ are produced respectively by MLPs consisting of two fully-connected (FC) layers, where the first FC layer is followed by the PReLU activation. The anchor rotation is predicted as Euler angles by the network and then is converted to a rotation matrix. The initial state $\mathbf{h}_0$ of each GRU layer is sampled from a normal distribution with mean zero and standard deviation 0.1. At inference, the initial state is set to zero and the model can be applied to motion sequences of arbitrary length.

\begin{figure}[t]
	\centering
	\includegraphics[width=8.0cm]{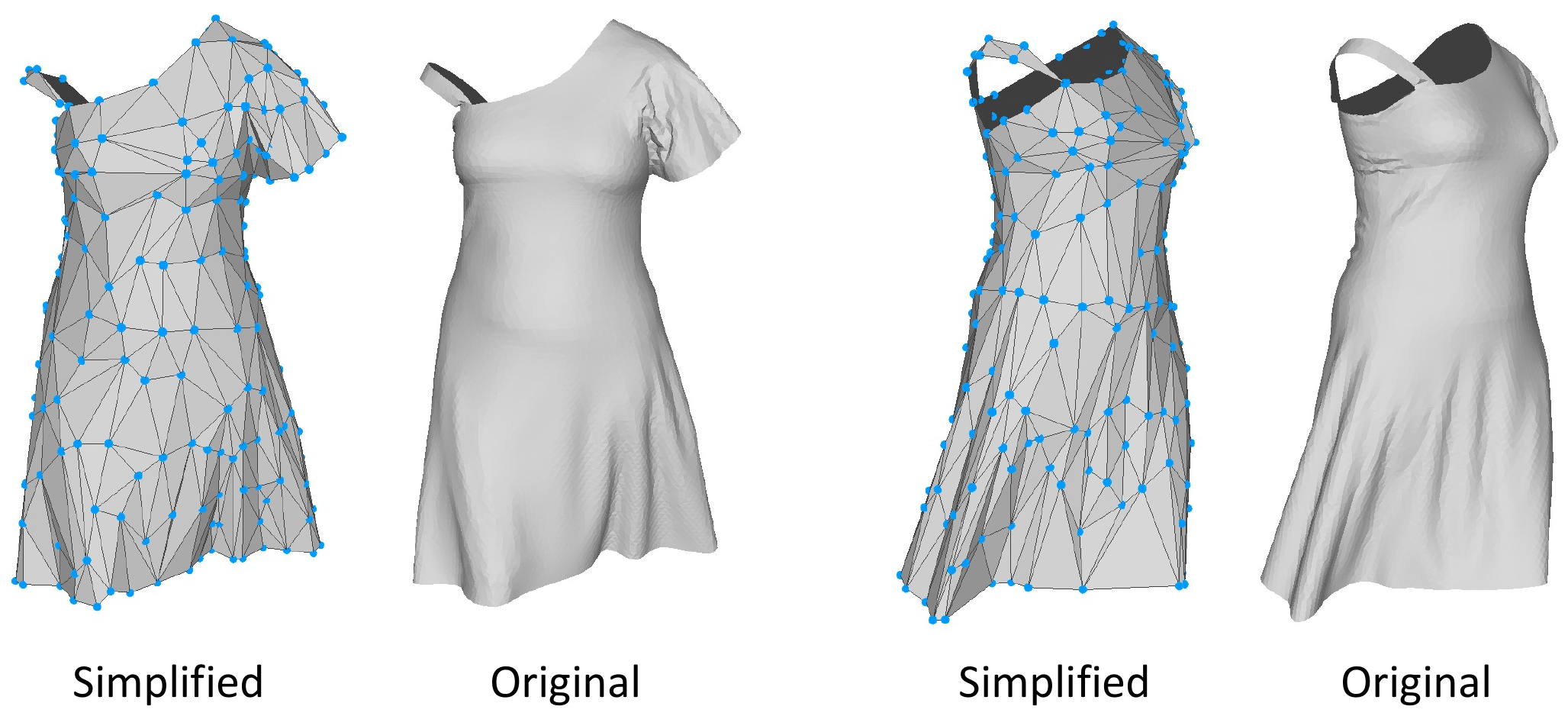}
	\vspace{-5pt}
	\caption{Visualization of simplified meshes used as the supervision for the adaptive anchor updating.}\label{fig_smp}
	\vspace{-10pt}
\end{figure}

Meshes simplified by Quadric Error Metric (QEM)~\cite{garland1997surface} are illustrated in Fig.~\ref{fig_smp}, which are used as the supervision for the adaptive anchor updating. These meshes preserve intrinsic geometric structures of the surface, \eg, folds and boundaries, which provide key topology information to learn representative anchors.

All compared models are trained using the hyper-parameters described in their papers. For TailorNet~\cite{patel2020tailornet}, we use the training code provided by its authors. VirtualBones~\cite{pan2022predicting} only releases the inference code, thus we re-implement the training code according to its inference code.

\begin{figure}[t]
	\centering
	\includegraphics[width=8.0cm]{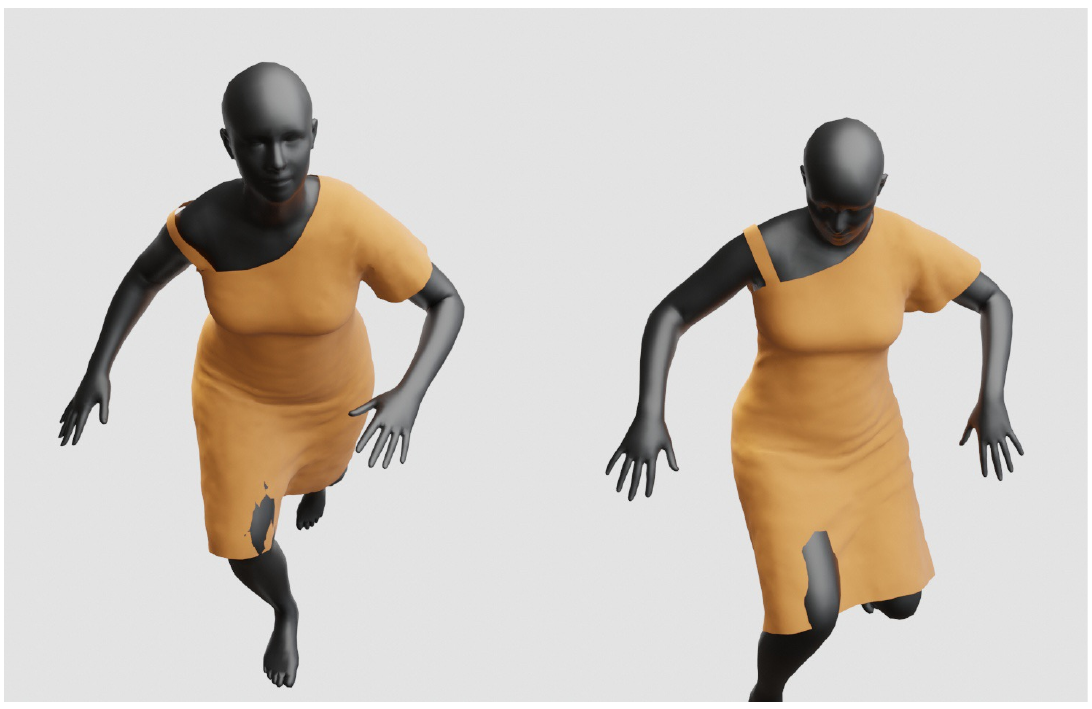}
	\caption{Failure cases for some extreme poses where garment-body interpenetration may appear.}\label{fig_fail}
\end{figure}

\section*{B. More Results}
\label{results}

Please refer to \url{https://semanticdh.github.io/AnchorDEF} for examples of our AnchorDEF for dynamic garment deformation in motion and qualitative comparison of our AnchorDEF with other 3D garment deformation methods~\cite{patel2020tailornet,pan2022predicting}. As shown in the demo video, given a body motion sequence which is unseen during training, our method can produce natural and realistic clothing dynamics and the garment deformation closer to the ground truth compared with other methods.

Some failure cases are shown in Fig.~\ref{fig_fail}. Garment-body interpenetration may appear for some extreme poses. Preventing or reducing such garment-body interpenetration is a future research direction.

\end{document}